\documentclass{IEEEmce}

\usepackage[colorlinks,urlcolor=blue,linkcolor=blue,citecolor=blue]{hyperref}
\usepackage{dblfloatfix}
\usepackage{float}
\usepackage{booktabs}
\usepackage{amsmath}
\usepackage{amssymb}
\usepackage{subcaption}
\usepackage{multirow}
\usepackage{makecell}
\usepackage{diagbox}
\usepackage{ulem}
\usepackage{xcolor}

\usepackage{hyperref}
\hypersetup{ 
 colorlinks=true, 
 linkcolor=red, 
 filecolor=blue, 
 citecolor = blue, 
 urlcolor=blue, 
 } 
 
\usepackage{upmath}

\jvol{XX}
\jnum{XX}
\paper{XX}
\jmonth{xxx/xxx}
\publisheddate{DD MM YYYY}
\currentdate{DD MM YYYY}
\jname{IEEE Consumer Electronics Magazine}
\pubyear{YYYY}
\doiinfo{MCE.YYYY.Doi Number}

\begin{document}

\sptitle{Running Head Title} 


\title{The Potential of Haptic Foundation Models}


\author{Jianquan Wang}


\affil{University of Ottawa}

\author{Haiwei Dong}
\affil{University of Ottawa and Dayan Technologies}

\author{Abdulmotaleb El Saddik}
\affil{University of Ottawa}


\markboth{Running Head Title}{Article Title}


\begin{abstract} Despite the success of foundation models in language and vision, their expansion into embodied AI is bottlenecked by a lack of generalized touch sensing. This limitation is especially relevant to consumer electronics, where smartphones, wearables, VR controllers, home robots, and health monitoring devices require safe and adaptive physical interaction. Constrained by hardware heterogeneity and the necessity of active physical data collection, current haptic models remain rigidly task-specific. To overcome these limitations, this article explores the transformative potential and developmental trajectory of Haptic Foundation Models (HFMs). We detail the paradigm shift required to transition from passive Large Language Models and Vision Language Models into active HFMs across four core dimensions: action coupling, physical dynamical representation space, continuous time-series data granularity, and action-conditioned future state prediction. Furthermore, we synthesize existing large-scale tactile datasets and benchmark UniTouch, AnyTouch, T3, and Sparsh on TacBench for force estimation, slip detection, and relative pose estimation.\end{abstract}

\maketitle

\enlargethispage{10pt}

\chapterinitial{Foundation Models} have become dominant in natural language processing and computer vision by learning from massive Internet-scale data and showing strong generalization capabilities. However, most existing foundation models still lack tactile perception, which is essential for understanding softness, friction, temperature, elasticity, and fine surface properties. As multimodal generative AI accelerates the integration of embodied agents into domestic environments \cite{10480226}, robust haptic perception has become increasingly important for consumer electronics. Haptic Foundation Models (HFMs) address this gap by serving as a tactile sensing and reasoning layer for user-facing devices, including smartphones and wearables, gaming peripherals and VR controllers, smart home robots, and home health monitors. In these scenarios, HFMs can interpret pressure, texture, grip stability, and contact patterns to support richer touch interfaces, realistic force feedback, safer grasping, assistive sensing, and contact-aware monitoring.

Building HFMs remains challenging because tactile data must be collected through active physical interaction and also because tactile sensors are highly heterogeneous. GelSight sensors \cite{yuan2017gelsight}, Hall-effect sensors, piezoelectric sensors, force-sensitive resistors, and other tactile devices produce signals with different dimensions, sampling rates, spatial resolutions, and physical meanings. As a result, tactile datasets collected from different sensors are difficult to merge directly, and current haptic models are often limited to dedicated sensors, datasets, and tasks. To address these limitations, HFMs aim to learn a unified representation space for heterogeneous tactile inputs, align low-level physical measurements with high-level semantic concepts, and support cross-task generalization and physical-state prediction.

Despite numerous challenges, multiple techniques are emerging to facilitate {the development of} HFMs. The primary enabler is the development of high-resolution vision-based tactile sensors, which serve as a critical bridge. By converting contact geometry into image-based representations, they allow tactile learning to directly leverage the massive pretrained capabilities of modern computer vision backbones, bypassing the need to design architectures for sparse signal arrays. High-fidelity simulation platforms enable the synthesis of large-scale tactile data, overcoming hardware bottlenecks in data collection. Additionally, contrastive learning in multimodality allows tactile embeddings to align with other modalities such as language and vision. The convergence of these technologies {has helped establish the technical foundations for further progress in embodied AI.}

However, the availability of tools does not automatically equate to intelligence. Unlike the standardized tokenization in language or pixel grids in vision, the path to a universal haptic model remains unclear due to ambiguous definitions and architectural fragmentation. {To move from specialized sensors toward more generalizable physical understanding, we first need to deconstruct the necessary components of such a system.} This leads us to a fundamental question that precedes all applications: \textit{What is a Haptic Foundation Model?}

\section{DEFINITION OF A HAPTIC FOUNDATION MODEL}
\begin{figure}[!ht]
\centerline{\includegraphics[width=\linewidth]{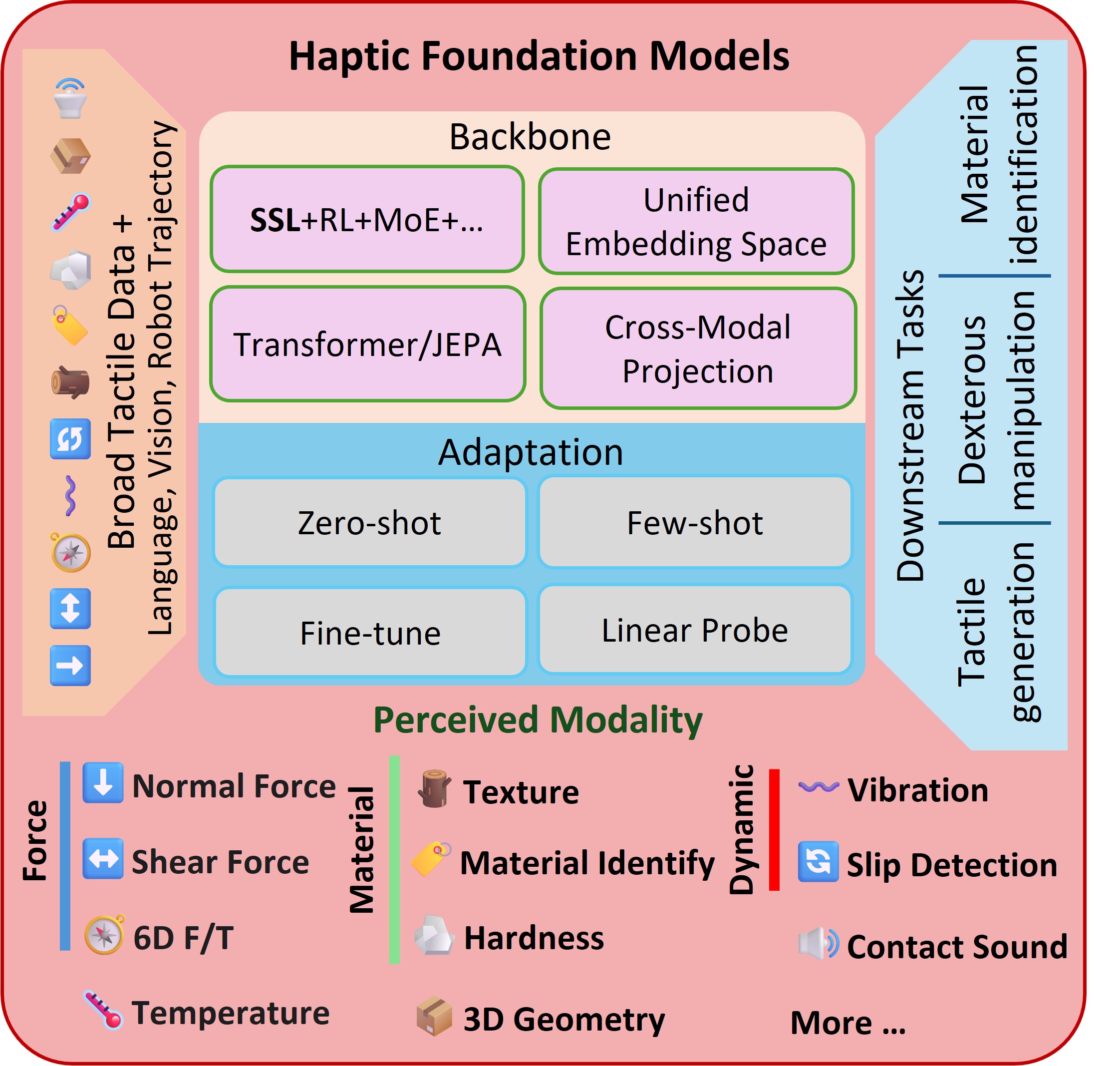}}
\caption{{Conceptual overview of a Haptic Foundation Model.} The model can process {multimodal data from heterogeneous tactile sensors and map them into a unified embedding space.} After proper adaptation, the model supports a {broad range of} downstream tasks. MoE: mixture of experts; RL: reinforcement learning; SSL: self-supervised learning.}
\label{fig:1}
\end{figure}

\begin{figure*}[htbp]
\vspace*{-25pt}
\centerline{\includegraphics[width=40pc]{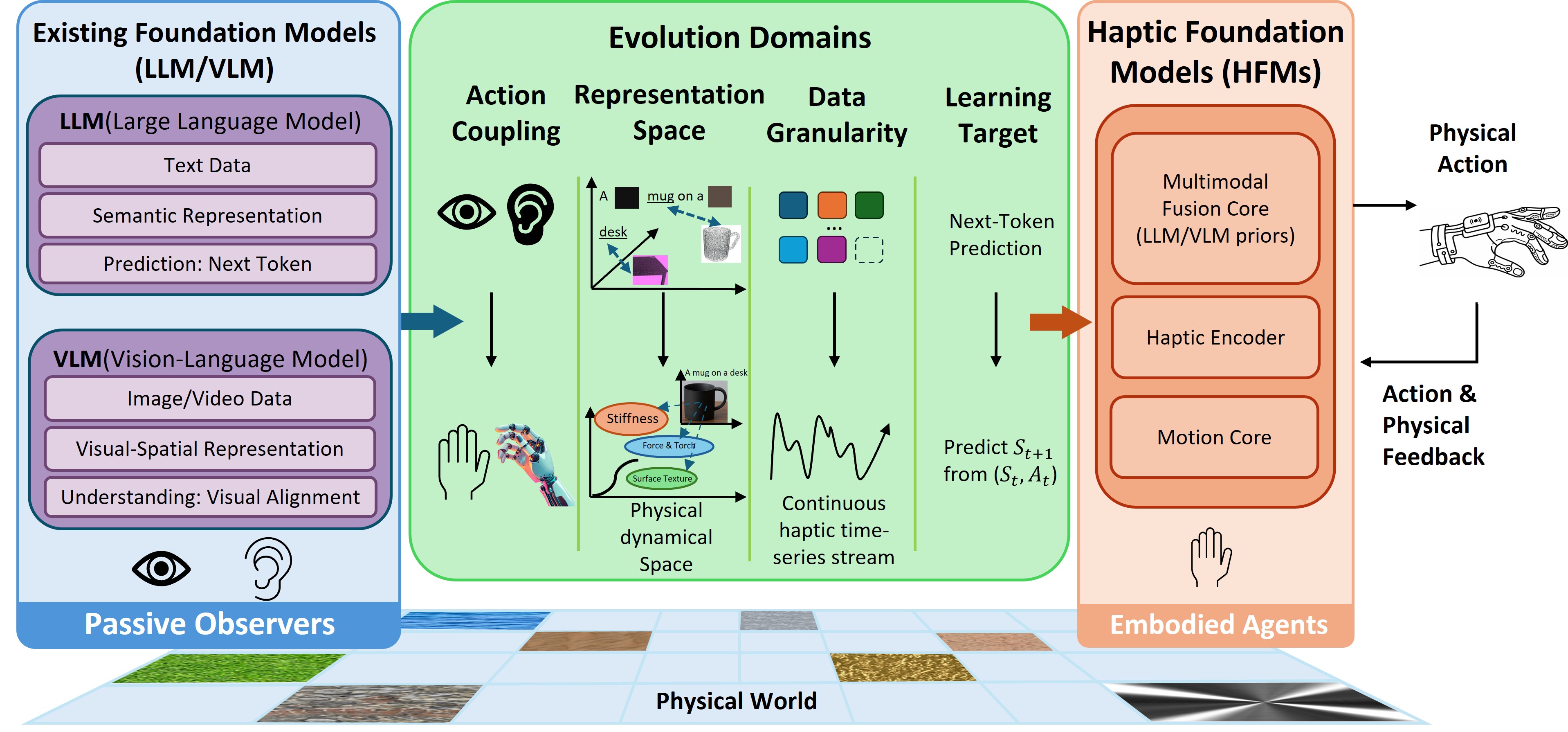}}
\caption{The evolution of HFMs from LLMs/VLMs. The framework illustrates the fundamental paradigm shift from passive observation to active embodied interaction across four key domains: action coupling, representation space, data granularity, and predictive learning targets.}
\label{fig:2}
\end{figure*}
An HFM inherits from the Foundation Model concept, {which is characterized by broad data coverage}, self-supervised pretraining, and downstream adaptability, as shown in Figure \ref{fig:1}. {The following discussion presents a prospective HFM framework and its possible functional components. These capabilities should be understood as design objectives and research directions rather than functions directly demonstrated by the benchmark later in this article.} Broad data encompasses diverse modalities, including high-resolution tactile images from vision-based sensors, time-series 6-dimensional force and torque signals, vibration signals, slip detection, thermal feedback, hardness, contact sound, touch video and trajectory, and more possible developing modalities. Unlike traditional task-specific models on narrow datasets, an HFM is trained on massive-scale, multi-source tactile observations, often integrated with cross-modal data {such as} vision and language to capture the physical {properties and interaction dynamics}. The pretraining phase typically employs self-supervised objectives, such as Masked Autoencoders (MAE) or cross-modal contrastive learning. This process allows the model to learn robust tactile representations that encode fundamental physical properties without the need for exhaustive manual labelling. Advanced methodologies from Large Language Models (LLMs) and Vision Language Models (VLMs) {may also inform HFM development.} For instance, Mixture of Experts (MoE) can be employed during the pretraining phase to scale parameters efficiently, while large-scale Reinforcement Learning (RL) can be utilized during the post-training alignment phase. {Once pretrained, an HFM could potentially support adaptation across a range of downstream robotic and VR tasks through fine-tuning or other lightweight adaptation strategies.} {Larger-scale pretraining may support broader transfer to novel objects or sensor configurations, although such generalization remains to be systematically established.}

\section{EVOLUTION FROM LLM/VLM TO HFM}
While LLMs and VLMs have demonstrated unprecedented capabilities in constructing high-dimensional semantic spaces and understanding visual geometry, they {primarily operate as} passive observers of {digital inputs}. As illustrated on the left side of Figure \ref{fig:2}, these models excel at processing pre-segmented, discrete streams of text or pixels, yet they lack the physical grounding required to comprehend implicit material properties such as mass, friction, stiffness, and tactile deformation. The evolution of HFMs, therefore, is not merely the addition of a new sensory modality, but a profound paradigm shift towards embodied AI. This evolution necessitates a structural and mathematical derivation across four core dimensions, as depicted in the central derivation path. The derivation of a Haptic Foundation Model (HFM) requires a fundamental shift from passive reception to active perception. Unlike vision or audition, haptics is intrinsically bidirectional. Physical properties cannot be perceived without initiating a motor action. To accommodate this, the development of an HFM {can be organized around four core dimensions}:

\begin{itemize}
    \item \textbf{Action-Coupled Active Perception:} An HFM {may} transition from a decoupled observer into an active agent. Because tactile exploration requires movement, sensory encoding {may need to be closely} integrated with motor control.
    
    \item \textbf{Physical Dynamical Representation Space:} This active character forces a transformation in the model's representation space. While VLMs map the world into a visual-spatial latent space, HFMs {might} anchor these visual priors into a physical dynamical space. Through cross-modal alignment, static visual geometry is translated into dynamic, force-aware representations that {capture force-dependent interaction dynamics}.
    
    \item \textbf{Continuous Spatiotemporal Data Granularity:} Bridging the physical gap requires redefining tokenization. While LLMs and VLMs rely on discrete tokens (e.g., BPE) or image patches, tactile feedback often consists of high-frequency, continuous time-series streams (e.g., force-torque or vibrotactile signals). For continuous time-series data, such as a continuous signal from a six-axis force-torque sensor, instead of discrete categorization, the continuous signal matrix $F \in \mathbb{R}^{T \times 6}$ is partitioned into overlapping temporal windows. These continuous windows are subsequently processed through {one-dimensional} convolutional layers to extract local temporal features. This operation maps the physical amplitudes and frequencies into a continuous embedding vector $E_{t} \in \mathbb{R}^{D}$. This physical tokenization {is intended to preserve} transient physical dynamics of contact, representing a fundamental structural difference from discrete linguistic tokens and enabling the model to process continuous physical measurements.
    
    \item \textbf{Action-Conditioned Future State Prediction:} These architectural shifts lead to a new learning target. The standard autoregressive objective of predicting the next discrete token is insufficient for physical interaction. Instead, as shown in the final transition of Figure \ref{fig:2}, the HFM {might evolve to model world dynamics by forecasting future haptic states from the current state and executed action.} For example, while a standard language model predicts the subsequent word in a textual sequence, a Haptic Foundation Model utilizes the current tactile state and a specific motor command, such as a robotic gripper increasing its grasp aperture by two millimeters, to forecast {changes in the subsequent continuous force-feedback signal.} {Action-conditioned forecasting could encourage the model to capture the physical consequences of interactions and may provide a basis for future closed-loop control.}
\end{itemize}

{In a prospective implementation, such an HFM could function as part of a closed-loop embodied dynamics model.} Through continuous physical action and real-time sensory feedback, {such a system could connect perception and actuation and may contribute to physically grounded robotic agents.}

\begin{table*}[!t]
\centering
\caption{Representative haptic datasets for training Haptic Foundation Models}
\label{table:dataset}
\scriptsize
\setlength{\tabcolsep}{3pt}
\renewcommand{\arraystretch}{1.08}
\begin{tabular}{p{4.8em}p{12em}p{9em}p{10em}p{13em}}
\hline
\textbf{Dataset} & \textbf{Modalities / Sensor Type} & \textbf{Scale} & \textbf{Key Annotations} & \textbf{HFM pretraining Suitability} \\
\hline
CLAMP \cite{thakkar2025clamp} &
Force, thermal, vision, audio; microphone, contact microphone, camera, thermal sensor, FSR, IMU &
12.3M data points; 5357 daily objects &
Compliance; material labels &
High: in-the-wild multimodal physical grounding \\

FoTa \cite{zhao2024transferable} &
Vision, tactile; 13 optical visuo-tactile sensors &
3.08M images &
Task data; spatial alignment &
Very high: unified format and cross-sensor alignment \\

FreeTacMan \cite{wu2025freetacman} &
Vision, tactile, proprioception, action; visuo-tactile sensors and wrist camera &
$>$3M pairs; 10k trajectories; 50 tasks &
6D pose; action instructions &
High: robot-free collection for manipulation dynamics \\

ToucHD \cite{feng2026anytouch2generaloptical} &
Vision, tactile, physical mechanics, action; force and visuo-tactile sensors &
2.42M samples; 1043 3D objects &
Dynamic action attributes; force parameters &
Very high: dynamic touch-force paired learning \\

exUMI \cite{xu2025exumi} &
Vision, tactile, proprioception; visuo-tactile sensors and camera &
$>$1M frames; diverse real-world scenes &
Action trajectories; contact states &
High: action-aware and portable collection \\

Touch100k \cite{cheng2024touch100klargescaletouchlanguagevisiondataset} &
Vision, tactile, language; visuo-tactile sensors &
100k triplets &
Multi-granularity text semantics &
High: tactile-language-vision semantic grounding \\

TacQuad \cite{fenganytouch} &
Vision, tactile, language; quad-aligned visuo-tactile sensors &
72,606 frames; 99 objects &
Cross-sensor spatial consistency &
Very high: cross-sensor matching and multimodal alignment \\

WIYH \cite{robotics2025world} &
Vision, tactile, language, action; tactile sensor, fisheye camera, IMU, IR camera, pinhole camera &
$>$1,000 hours; 100 skills &
First-person daily scenarios &
High: egocentric, long-horizon, human-centric manipulation \\
\hline
\end{tabular}
\vspace*{-8pt}
\end{table*}

\begin{figure}[t]
\vspace*{-5pt}
\centerline{\includegraphics[width=17pc]{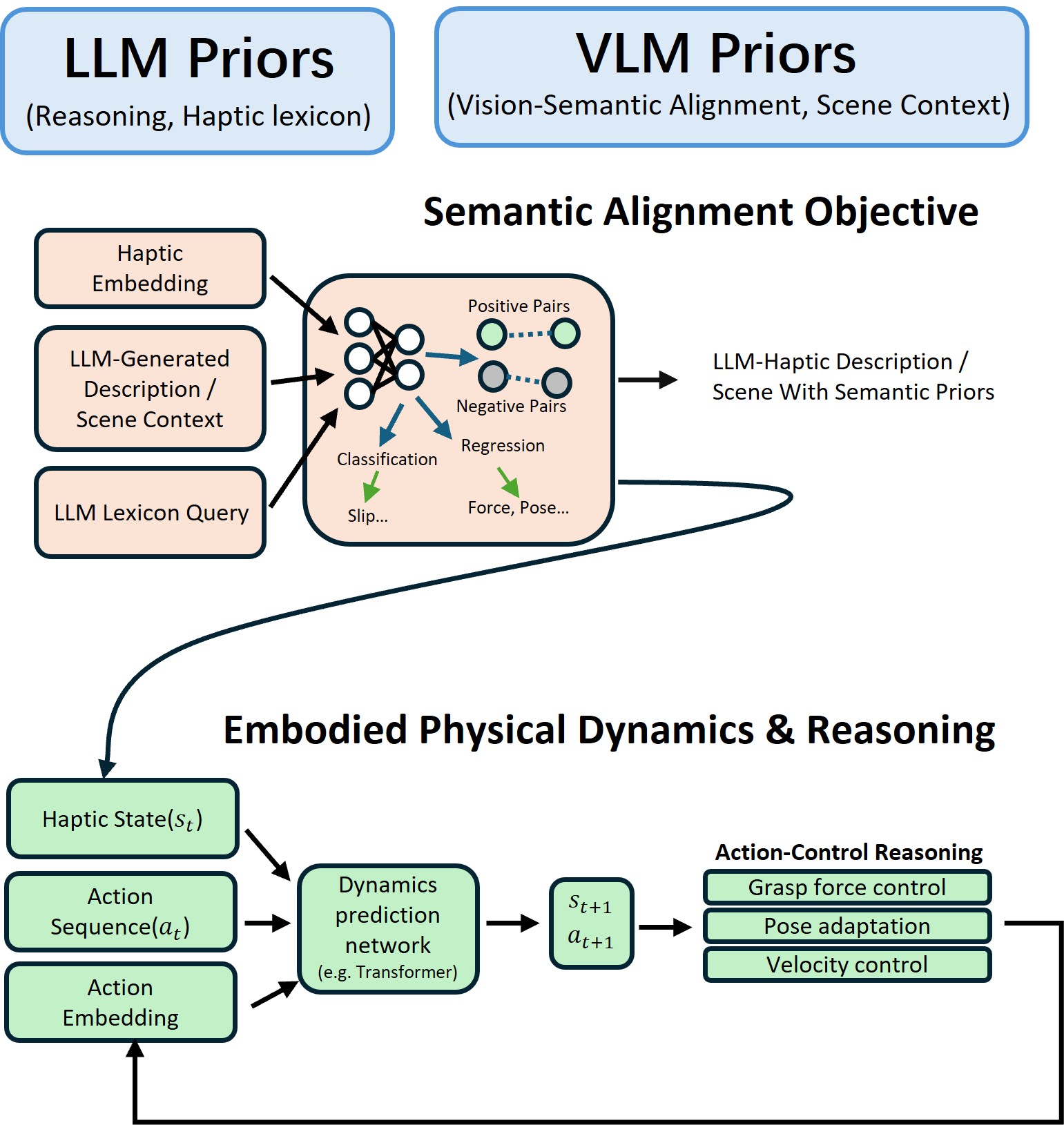}}
\caption{{Prospective} architectural framework for evolving HFM. The framework integrates a semantic alignment objective to distill LLM/VLM priors into haptic embeddings, which are subsequently fused with embodied physical dynamics {with the goal of supporting action-aware tactile reasoning and future closed-loop control} (e.g., grasp force, pose, and velocity).}
\label{fig:3}
\end{figure}

The proposed technical roadmap for evolving a HFM from general-purpose LLMs and VLMs is illustrated in Figure \ref{fig:3}. This evolution process is structured into two interconnected optimization layers. In the upper semantic alignment stage, raw haptic embeddings are regularized against LLM-generated descriptions and VLM scene contexts through contrastive learning and dictionary-based attribute queries. {This step is intended to encourage the resulting haptic representations to encode physical semantics, such as stiffness, texture, or slip probability.} As indicated by the integration path, {these semantically enriched embeddings are then used as the haptic state input to the embodied dynamics module.} By combining enriched tactile features with action sequences and state histories, a transformer-based dynamics network predicts future tactile states. {The predicted tactile states could be used to inform control-relevant quantities such as grasp force, endpoint pose, or velocity limits.} 

\section{KEY TECHNOLOGIES IN HFM}

Following these conceptual shifts, the development of HFMs depends on several enabling technologies that {are intended to support} robust sensing, heterogeneous data alignment, and dynamic physical prediction.
Unlike visual or linguistic inputs, physical tactile sensors, when deployed in consumer electronics or robotic interfaces, experience continuous mechanical wear, aging, and frequent replacement. This {can lead to gradual} variations in {internal sensor characteristics} that, if unaddressed, ultimately result in significant perception errors. Consequently, adaptive self-calibration has emerged as a critical resilience mechanism. Utilizing self-supervised temporal prediction during active touch actions, models can dynamically recalibrate their learnable encoders without manual intervention \cite{bhirangi2021reskin}. {Such adaptation may improve long-term operational reliability and help realign changing sensor signals within a shared representation space.}

Beyond managing hardware heterogeneity, the architectural handling of data heterogeneity presents another pivotal challenge. Tactile perception inherently involves the integration of asymmetric data streams, including low-bandwidth, sparse physical signals (e.g., force, thermal, vibration) and high-bandwidth visuo-tactile or external video modalities. A major risk in such hybrid systems is that sparse tactile features might be submerged by the overwhelming data density of standard fully-connected Transformer architectures. {As one illustrative implementation of multimodal fution, we consider} bottleneck fusion tokens with cross-attention mechanisms \cite{higuera2025tactile}. {This choice is used to explain how tactile information could interact with other modalities and is not intended as a universal or mandatory design for HFMs}. To construct the unified embedding space and handle vastly different data dimensions from heterogeneous sensors, {In this illustrative design, modality-specific tokenization strategies and projection layers are used.} For instance, {high-resolution visuo-tactile} data from GelSight sensors, represented as an image tensor $I \in \mathbb{R}^{H \times W \times C}$, is divided and embedded into patch tokens $E_{v}$. Conversely, {low-dimensional} continuous time series data, such as a {six-dimensional force-torque} signal $F \in \mathbb{R}^{T \times 6}$, is processed through {one-dimensional} convolutional layers to generate temporal embeddings $E_{f}$. To align these varying dimensions into a shared latent space $\mathbb{R}^{D}$, specialized linear projection layers $P_{v}$ and $P_{f}$ are applied to produce unified tokens $Z_{v} = P_{v}(E_{v})$ and $Z_{f} = P_{f}(E_{f})$. Furthermore, to prevent the sparse tactile features from being dominated by the high density visual data, the {illustrative} architecture utilizes a {cross-attention} mechanism with learnable bottleneck fusion tokens $B$. The fusion process is mathematically expressed as
$$Z_{fused} = \text{Attention}(B, [Z_{v}, Z_{f}], [Z_{v}, Z_{f}])$$
{ This mechanism is intended to regulate information exchange between sparse tactile signals and higher-dimensional visual inputs, although its effectiveness relative to alternative fusion strategies requires empirical evaluation. Other fusion strategies, including early feature fusion, cross-attention, hierarchical fusion, mixture-of-experts routing, and late decision fusion, may also be appropriate depending on the sensor configuration, computational budget, and downstream task. Determining the most effective fusion architecture for HFMs remains an open research question.}

Finally, while managing input diversity is vital, the learning target of the HFM {may} extend beyond localized, two-dimensional contact patches to comprehend true, dynamic three-dimensional physical interactions. To this end,{recent work has explored} pyramid feature learning structures to capture {multiscale temporal features} across continuous time-series data \cite{feng2026anytouch2generaloptical}. By fusing these tactile temporal sequences with proprioceptive arm movements or camera inputs to infer 3D structural geometries and spatial scene relationships, {such fusion may support spatiotemporal representations that connect local contact information with broader physical context.}

\section{HAPTIC DATASETS}

As embodied AI advances toward contact-rich manipulation, relying solely on vision and language modalities reveals significant limitations. {Addressing substantial} hardware heterogeneity of tactile sensors requires a paradigm shift from small, task-specific benchmarks to million-scale pretraining corpora and high-density multimodal alignment datasets. Table \ref{table:dataset} summarizes the representative large-scale datasets that {contribute to the emerging data infrastructure for HFMs}. 

These datasets reveal three main trajectories for HFM data construction. First, datasets such as CLAMP, ToucHD, and FoTa increase data scale and sensor diversity, supporting multimodal physical grounding and cross-sensor alignment. Second, FreeTacMan and exUMI connect tactile feedback with action, pose, and proprioception, {thereby supporting research on contact-rich manipulation.} Third, Touch100k, TacQuad, and WIYH emphasize tactile-language-vision alignment and human-centric daily scenarios, helping translate localized contact signals into semantic and consumer-relevant representations.

As physical AI {is increasingly} integrated into everyday devices \cite{dong2026physical}, representative systems such as FreeTacMan, exUMI, and CLAMP illustrate two complementary directions for HFM data collection{:} action-coupled visuo-tactile manipulation and multimodal physical sensing. FreeTacMan and exUMI emphasize portable, action-aware collection with tactile sensing, proprioception, and visual context, whereas CLAMP highlights multi-physics sensing through force, thermal, audio, vision, and inertial signals. Together, these systems show that future HFM datasets {may} capture both manipulation dynamics and rich physical properties in consumer-relevant environments.

While extant datasets establish robust baselines, future dataset curation {may} evolve to drive the next generation of embodied intelligence. First, to eliminate the mechanical latency of traditional teleoperation, data acquisition is shifting toward robot-free wearable systems (e.g., consumer-grade haptic gloves and smartwatches). These devices can capture unstructured, human-level skills in daily scenarios with {while preserving human motion and proprioceptive context}. Second, {given the high cost of physical data collection, real-world datasets may increasingly be complemented by high-fidelity simulation and domain randomization. Such simulation could expand dataset scale and support sim-to-real studies. Combining simulated data with wearable and crowdsourced observations may ultimately facilitate the development of more predictive tactile models.}

Although recent haptic datasets have reached millions of frames, the lack of standardized data formats remains a major bottleneck for Haptic Foundation Models. Camera-based tactile sensors differ in form factor, camera configuration, illumination design, resolution, and signal meaning, while other modalities such as force-sensitive resistors or thermal sensors produce one-dimensional time-series data with different physical units. Without aligned representations or a shared physical coordinate framework, signals collected from different sensors cannot be directly compared or merged into a unified training corpus. Therefore, standardized data representation across heterogeneous tactile sensors is as important as dataset scale for enabling massive HFM pretraining.

\begin{table*}[htbp]
\centering
\caption{Model Attribute Comparison for Representative Haptic Foundation Models}
\label{tab:model_attributes}
\resizebox{\textwidth}{!}{
\begin{tabular}{p{2cm}p{2.6cm}p{3.2cm}p{3cm}ccp{2cm}}
\toprule
\textbf{Model} & \textbf{pretraining Data} & \textbf{Supported Sensors} & \textbf{Input Modalities} & \textbf{Parameters} & \textbf{{Estimated} Comp. Cost} &\textbf{Eval. Protocol} \\ 
\midrule
\textbf{UniTouch {\cite{huggingfaceChfengTouchLLMMain}}} & Vision-tactile aligned pairs & DIGIT, GelSight, Taxim, Tacto & Vision, Tactile & \textasciitilde 86 M & Moderate & Frozen encoder + linear head \\ 
\textbf{AnyTouch {\cite{githubGitHubGeWuLabAnyTouch}}} & Cross-sensor static \& dynamic tactile data & Multiple visuo-tactile sensors & Vision, Tactile, Temporal & \textasciitilde 86 M & Moderate& Frozen encoder + linear head \\ 
\textbf{T3-small {\cite{huggingfaceAlanzmitFoundationTactileMain}}} & Multi-sensor task-agnostic data (FoTa) & DIGIT, GelSight & Vision, Tactile & 45 M & Low& Frozen encoder + linear head  \\ 
\textbf{T3-medium {\cite{huggingfaceAlanzmitFoundationTactileMain}}} & Multi-sensor task-agnostic data (FoTa) & DIGIT, GelSight & Vision, Tactile & 174 M & Moderate& Frozen encoder + linear head \\ 
\textbf{T3-large {\cite{huggingfaceAlanzmitFoundationTactileMain}}} & Multi-sensor task-agnostic data (FoTa) & DIGIT, GelSight & Vision, Tactile & 304 M & High& Frozen encoder + linear head \\ 
\textbf{Sparsh-MAE {\cite{huggingfaceFacebooksparshmaebaseMain}}} & 460k+ unlabeled tactile images & DIGIT, GelSight, GelSight Mini & Vision, Tactile & \textasciitilde 86 M & Moderate& Frozen encoder + linear head \\ 
\textbf{Sparsh-DINO {\cite{huggingfaceFacebooksparshdinobaseMain}}} & 460k+ unlabeled tactile images & DIGIT, GelSight, GelSight Mini & Vision, Tactile & \textasciitilde 86 M & Moderate& Frozen encoder + linear head \\ 
\textbf{Sparsh-DINOv2 {\cite{huggingfaceFacebooksparshdinov2baseMain}}} & 460k+ unlabeled tactile images & DIGIT, GelSight, GelSight Mini & Vision, Tactile & \textasciitilde 86 M & Moderate& Frozen encoder + linear head\\ 
\textbf{Sparsh-IJEPA {\cite{huggingfaceFacebooksparshijepabaseMain}}} & 460k+ unlabeled tactile images & DIGIT, GelSight, GelSight Mini & Vision, Tactile & \textasciitilde 86 M & Moderate& Frozen encoder + linear head \\ 
\textbf{Sparsh-VJEPA {\cite{huggingfaceFacebooksparshvjepabaseMain}}} & 460k+ unlabeled tactile images & DIGIT, GelSight, GelSight Mini & Vision, Tactile, Temporal & \textasciitilde 86 M & Moderate & Frozen encoder + linear head\\ 
\bottomrule
\multicolumn{7}{p{17.5cm}}{\footnotesize * Note: Computational costs are qualitative categories estimated from parameter scale and architectural complexity, rather than hardware-independent latency measurements. All benchmark results reported in the following performance table use the same frozen-encoder linear-probe protocol: the pretrained backbone is fixed, and only lightweight downstream heads are trained on the TacBench training split. Parameter sizes for UniTouch, AnyTouch, and Sparsh correspond to standard ViT-Base-scale backbone configurations commonly used in their released implementations.}\\
\end{tabular}
}
\end{table*}

\begin{table*}[htbp]

\begin{center}
    \centering
    \setlength{\tabcolsep}{2pt}
    \caption{{Frozen-encoder linear-probe benchmark on TacBench.}}
    \label{tab:performance_baseline}
    \resizebox{\textwidth}{!}{
    \begin{tabular}{lcccccc}
        \toprule
        \diagbox{Model}{Task} & \makecell{Force\\-DIGIT} &\makecell{Force\\-GelSight}& \makecell{Slip\\-DIGIT} &\makecell{Slip\\-GelSight}& {\makecell{Pose\\Trans.}} 
        &{\makecell{Pose\\Rot.}}
        \\
        \midrule
UniTouch
& $815.0 \pm 0.4$
& $296.9 \pm 0.2$
& $0.463 \pm 0.002$
& $0.468 \pm 0.006$
& $4.397 \pm 0.014$
& $1.780 \pm 0.089$
\\

AnyTouch
& $826.4 \pm 14.0$
& $284.0 \pm 0.8$
& $0.470 \pm 0.006$
& $\mathbf{0.499 \pm 0.004}$
& $4.412 \pm 0.007$
& $2.845 \pm 0.690$
\\

T3-small
& $797.4 \pm 1.0$
& $334.4 \pm 3.0$
& $0.455 \pm 0.004$
& $0.448 \pm 0.010$
& $4.425 \pm 0.006$
& $1.862 \pm 0.208$
\\

T3-medium
& $\mathbf{701.8 \pm 14.0}$
& $322.8 \pm 5.2$
& $0.446 \pm 0.013$
& $0.447 \pm 0.015$
& $4.400 \pm 0.010$
& $2.446 \pm 1.143$
\\

T3-large
& $800.4 \pm 1.6$
& $\mathbf{275.9 \pm 5.7}$
& $0.463 \pm 0.004$
& $0.461 \pm 0.007$
& $4.491 \pm 0.037$
& $5.885 \pm 0.961$
\\

Sparsh-MAE
& $716.4 \pm 0.3$
& $336.3 \pm 0.5$
& $0.448 \pm 0.006$
& $0.426 \pm 0.018$
& $4.389 \pm 0.011$
& $2.227 \pm 0.543$
\\

Sparsh-DINO
& $744.0 \pm 2.8$
& $336.7 \pm 0.9$
& $0.456 \pm 0.012$
& $0.432 \pm 0.016$
& $4.423 \pm 0.007$
& $2.216 \pm 0.385$
\\

Sparsh-DINOv2
& $766.7 \pm 0.3$
& $338.8 \pm 3.2$
& $0.444 \pm 0.006$
& $0.444 \pm 0.007$
& $4.436 \pm 0.009$
& $2.075 \pm 0.486$
\\

Sparsh-IJEPA
& $840.2 \pm 1.0$
& $349.2 \pm 0.8$
& $\mathbf{0.471 \pm 0.002}$
& $0.427 \pm 0.006$
& $\mathbf{4.382 \pm 0.005}$
& $\mathbf{1.674 \pm 0.071}$
\\

Sparsh-VJEPA
& $889.1 \pm 1.4$
& $353.3 \pm 0.9$
& $0.442 \pm 0.002$
& $0.438 \pm 0.012$
& $4.413 \pm 0.004$
& $1.697 \pm 0.084$
\\
        \midrule
        Metric & RMSE[mN]$\downarrow$&RMSE[mN]$\downarrow$&F1 score $\uparrow$&F1 score $\uparrow$&RMSE[mm]$\downarrow$&RMSE[$^\circ$]$\downarrow$\\
        \bottomrule
        \multicolumn{6}{p{12.5cm}}{\footnotesize Note 1: DIGIT and GelSight denote sensor-specific subsets in TacBench. Both are vision-based tactile sensing platforms, but they differ in sensor design, resolution, and contact-image characteristics.}\\
    \end{tabular}}
    \end{center}

\end{table*}

\section{BENCHMARKING HAPTIC FOUNDATION MODELS}

{To compare the representation quality of existing haptic foundation models under a controlled downstream setting, we evaluated ten pretrained encoder variants from four representative model families: UniTouch \cite{unitouch}, AnyTouch \cite{fenganytouch}, T3 \cite{zhao2024transferable}, and Sparsh \cite{higuera2024sparshselfsupervisedtouchrepresentations}. The benchmark is not intended to validate the full HFM framework described above. Instead, it compares how readily contact-force, slip-state, and relative-motion information can be linearly recovered from existing frozen tactile representations on TacBench.
All models were evaluated using a frozen-encoder linear-probe protocol. For each model, a released pretrained checkpoint was loaded, the encoder was placed in evaluation mode and its parameters were frozen, and only a task-specific linear readout head was trained. Each encoder retained its checkpoint-compatible preprocessing and fixed feature aggregation. The benchmark therefore measures the linear accessibility of task-relevant information in the pretrained representations rather than end-to-end adaptation performance. Force estimation and slip detection were evaluated separately on the DIGIT and GelSight subsets. Both tasks used single-frame tactile representations. Force estimation was formulated as linear regression of a three-dimensional force vector and evaluated using vector RMSE in millinewtons. Slip detection was formulated as framewise binary classification using a linear classifier trained with binary cross-entropy, with F1 score as the primary metric. Although slip is inherently dynamic, this protocol evaluates whether a single frozen tactile representation contains information that is linearly predictive of the slip label. Relative pose estimation was evaluated on the DIGIT pose subset using paired tactile frames at $t-5$ and $t$. Each frame was independently processed by the same frozen encoder, and the two resulting representations were concatenated before applying linear translation and rotation heads. The target was the sensor-relative planar motion $(\Delta x,\Delta y,\Delta\theta)$ extracted from the TacBench relative transformation matrices. Translation was evaluated using Euclidean RMSE in millimeters, while rotation was evaluated using wrapped angular RMSE in degrees. Because no trainable temporal module was used, this task evaluates the linear recoverability of relative-motion information from paired frozen representations rather than the native temporal modeling capability of each architecture.
Data were partitioned at the trajectory or bag level to prevent temporally adjacent samples from appearing in different splits. Model selection was based only on validation performance, and the test split was used for final evaluation. For every model--task combination, the pretrained encoder and data split were fixed, while the downstream linear head was trained independently using random seeds 0, 1, and 2. Results are reported as mean $\pm$ sample standard deviation across the three runs.
Table \ref{tab:model_attributes} summarizes the evaluated model variants, pretraining data, supported sensors, input formulations, and parameter scales. Because the models differ in architecture, pretraining objective, sensor coverage, and checkpoint-specific preprocessing, the results should be interpreted as a controlled comparison of frozen linear-probe representation quality rather than as evidence of architectural equivalence.}

{As shown in Table \ref{tab:performance_baseline}, the evaluated models exhibit complementary and sensor-dependent strengths, with no single frozen representation dominating all tasks. The T3 family includes the lowest-mean force-estimation results on both sensor subsets, whereas AnyTouch and Sparsh-IJEPA achieve the highest mean slip-detection scores on GelSight and DIGIT, respectively. Pose-translation errors are tightly clustered across models, while rotation estimation reveals clearer differences, with Sparsh-IJEPA, Sparsh-VJEPA, and UniTouch showing comparatively strong performance.
The relatively small standard deviations for most force, slip, and translation results suggest limited run-to-run variability under the evaluated downstream-head training settings. Some rotation results exhibit larger variability, indicating greater sensitivity to downstream-head initialization and optimization. Model scale does not produce a uniform improvement across tasks, indicating that representation quality depends more strongly on the interaction among pretraining strategy, sensor characteristics, and downstream physical quantity than on model size alone.
}

\section{CONCLUDING REMARKS AND SAFETY OUTLOOK}
{HFMs offer a promising direction for addressing the physical-grounding limitations of vision- and language-centered models.} By integrating large-scale pretraining, cross-modal alignment, and active perception, {HFMs offer substantial research potential.} {They may help mitigate sensor heterogeneity and improve the representation of contact-rich interactions.} Looking ahead, further breakthroughs are likely to {depend heavily} on the massive scaling of real-world multimodal datasets. Additionally, the deep integration of high-fidelity physical simulation engines {may play an important role in reducing the sim-to-real gap.}. More importantly, {closer coupling between perception and action may support the evolution of HFMs from passive state estimators toward predictive, closed-loop policies.} Ultimately, this interaction infrastructure could make HFMs {a useful component of future general-purpose robotic systems} in complex, real-world environments. 

Despite this potential, deploying HFMs in consumer-facing embodied systems raises important reliability, safety, and ethical challenges. {When integrated into} home robots, wearable devices, haptic gloves, or assistive monitors{, HFMs} may influence physical actions around humans, incorrect tactile judgments can lead to unsafe grasp forces, dropped or damaged objects, unstable manipulation, or inappropriate physical assistance. Key failure modes include sensor drift, aging, elastomer wear, calibration changes, out-of-distribution materials and contact patterns, environmental changes such as temperature and humidity, and sim-to-real gaps in friction, deformation, compliance, and sensor noise. Consumer haptic systems may also collect continuous contact and interaction traces in private environments, raising privacy and consent concerns.

These risks suggest that future HFMs should be developed with safety-aware deployment mechanisms, including continuous self-calibration, uncertainty estimation, out-of-distribution detection, conservative force and torque limits, mechanically compliant hardware, human override, privacy-preserving on-device processing, and systematic testing across sensor wear, material diversity, and environmental conditions. The transition from tactile recognition to predictive physical policies should therefore be accompanied by reliability verification and risk-aware control, so that HFMs can support safe embodied intelligence in human-centric consumer environments.

\bibliographystyle{IEEEtran}
\bibliography{ref}

\begin{IEEEbiography}{Jianquan Wang} is a PhD student at the University of Ottawa. He is a Student Member of IEEE and ACM. Contact him at jianquan.wang@uottawa.ca.
\end{IEEEbiography}

\begin{IEEEbiography}{Haiwei Dong} is an Adjunct Professor at the University of Ottawa and Chief Technology Officer at Dayan Technologies. He is a Senior Member of IEEE and ACM. Contact him at haiwei.dong@ieee.org.
\end{IEEEbiography}

\begin{IEEEbiography}{Abdulmotaleb El Saddik} is a Distinguished University Professor at the University of Ottawa. He is a Fellow of Royal Society of Canada, a Fellow of IEEE, an ACM Distinguished Scientist and a Fellow of the Engineering Institute of Canada and the Canadian Academy of Engineers. Contact him at elsaddik@uottawa.ca.
\end{IEEEbiography}

\end{document}